\documentclass{article} 
\PassOptionsToPackage{table,dvipsnames}{xcolor}
\usepackage{iclr2025_conference,times}

\usepackage{amsmath,amsfonts,bm}

\def\eqref#1{equation~\ref{#1}}

\def\1{\bm{1}}

\DeclareMathAlphabet{\mathsfit}{\encodingdefault}{\sfdefault}{m}{sl}
\SetMathAlphabet{\mathsfit}{bold}{\encodingdefault}{\sfdefault}{bx}{n}

\usepackage{hyperref}
\usepackage{url}
\usepackage{xspace}
\usepackage{booktabs} 
\usepackage{graphicx} 
\usepackage{wrapfig}
\usepackage{booktabs}
\usepackage{multirow}
\usepackage{multicol}
\usepackage{subcaption} 
\definecolor{lightpurple}{RGB}{230, 230, 255}
\definecolor{lightgreen}{RGB}{235, 255, 235}
\definecolor{veronica-red}{RGB}{196,30,58}

\title{\textbf{CODA}: \textbf{CO}ordinating the Cerebrum and Cerebellum for a \textbf{D}ual-Brain Computer Use \textbf{A}gent with Decoupled Reinforcement Learning.}

\author{Zeyi Sun$^{*1,2}$, Yuhang Cao$^{*2}$, Jianze Liang$^{*2}$, Qiushi Sun$^{*4}$, Ziyu Liu$^{*1,2}$ Zhixiong Zhang$^{1,2}$\\
\textbf{Yuhang Zang$^{\dagger2}$, Xiaoyi Dong$^{2,3}$, Kai Chen$^{2}$, Dahua Lin$^{2,3}$, Jiaqi Wang$^{\dagger2}$} \\
$^1$Shanghai Jiao Tong University \quad
$^2$Shanghai AI Laboratory \quad \\
$^3$The Chinese University of Hong Kong \quad
$^4$The University of Hong Kong
}

\newcommand{\methodname}{CODA\xspace}
\iclrfinalcopy 
\begin{document}

\maketitle

\renewcommand{\thefootnote}{\fnsymbol{footnote}}
{\let\thefootnote\relax\footnotetext{
\noindent \hspace{-5mm}$\dagger$ Corresponding Authors.
\noindent \hspace{-5mm}\quad \quad $^{*}$ Equal contribution \\
}   
}

\begin{abstract}


Autonomous agents for Graphical User Interfaces (GUIs) face significant challenges in specialized domains like scientific computing, require both long-horizon planning and precise, fine-grained execution. Existing approaches suffer from a trade-off: generalist agents excel at planning but falter in execution, while specialized agents show the opposite weakness. While recent compositional frameworks attempt to bridge this gap by combining a "planner" and an "actor," they are typically static and non-trainable, preventing adaptation from experience—a critical limitation given the scarcity of high-quality data in scientific domains.
To address these limitations, we introduce \textbf{CODA}, a novel and trainable compositional framework that synergizes a generalist planner (Cerebrum) with a specialist executor (Cerebellum), trained with a dedicated two-stage training pipeline. The first stage, \textbf{Specialization}, employs a decoupled GRPO approach to train an expert planner for each scientific application individually, bootstrapping from a small set of initial task trajectories. The second stage, \textbf{Generalization}, aggregates all successful trajectories from all specialized experts. This consolidated, high-quality dataset is then used to perform supervised fine-tuning (SFT) on the final planner, equipping it with the robust, cross-domain capabilities of a generalist.
Evaluated on four challenging applications from the ScienceBoard benchmark, our framework significantly outperforms the baseline and establishes a new state-of-the-art (SOTA) among open-source models. 
Our models and code are available at \url{https://github.com/OpenIXCLab/CODA}.
\end{abstract}
\section{Introduction}

Autonomous agents for Graphical User Interfaces (GUIs) ~\citep{claude,operator,qin2025uitars,lin2024showui,wu2024atlas,hong2023cogagent} promise to automate a wide range of digital tasks~\citep{zhou2023webarena,xie2024osworld}. However, their application in specialized domains such as scientific computing and engineering analysis remains highly challenging~\citep{sun2025scienceboard}. These environments pose two primary difficulties: first, their interfaces are highly complex, requiring precise and fine-grained actions; second, the problems they address are intrinsically complicated, demanding long-horizon planning to achieve effective solutions.

Effective agency for computer task automation in these domains requires both high-level planning and low-level execution as well as domain knowledge. However, current models exhibit a clear trade-off. Generalist models like Qwen2.5-VL~\citep{bai2025qwen2} provide robust planning capabilities but often struggle with the precise grounding needed for reliable execution. Conversely, specialized agents~\citep{wu2024atlas,wu2025gui,xie2025scaling} like UI-Tars~\citep{qin2025uitars} are highly proficient in execution, yet their capacity for complex, high-level planning is more constrained. 

To bridge this gap, a natural approach has been to develop compositional frameworks that explicitly decouple planning from execution, effectively pairing a generalist ``cerebrum" with a specialist ``cerebellum"~\citep{agashe2024agent,agashe2025agent}. 
While promising, these pioneering approaches are fundamentally limited. They are typically static and non-trainable, relying on powerful, often closed-source models as their core planner. This design introduces significant drawbacks: it compromises transparency and replicability, and most critically, prevents the agent from learning and adapting through experience. 

\begin{figure}[t]
    \centering
    \includegraphics[width=1.0\textwidth]{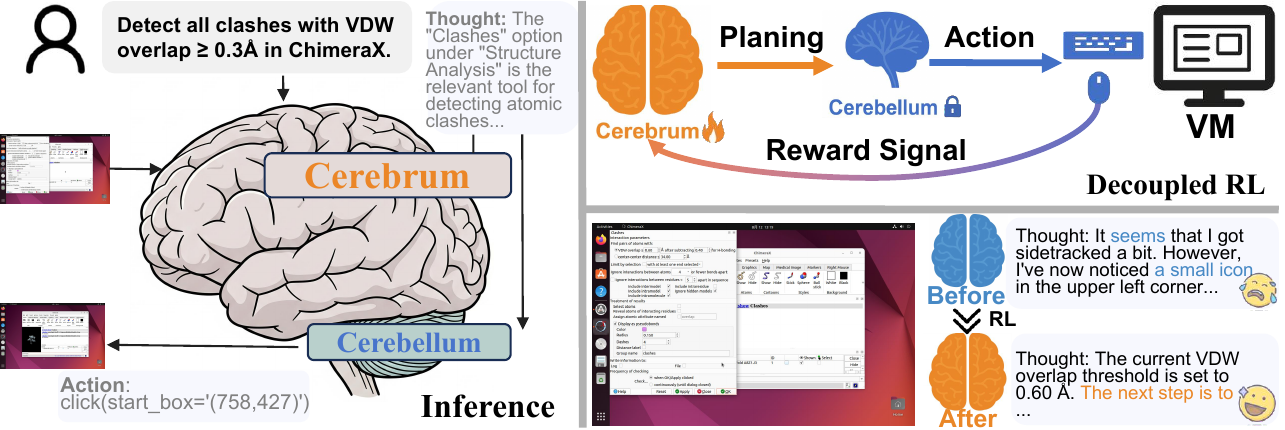}
    \caption{\textbf{Overall architecture of the proposed learnable Planner–Executor framework.} 
    Analogous to the relationship between the cerebrum and the cerebellum in the human brain, the Planner (cerebrum) generates high-level thoughts based on the history and screenshots, while the Executor (cerebellum) executes concrete GUI actions accordingly.}
    \label{fig:teaser}
\end{figure}

This architectural decoupling is not merely an engineering convenience but is inspired by the functional architecture of the human brain (illustrated in Fig.\ref{fig:teaser}). The specialization of high-level planning (the Cerebrum) and low-level motor control (the Cerebellum) is a key aspect of human intelligence. Crucially, these structures exhibit different learning patterns: the Cerebellum, once mature, provides stable and broadly applicable motor skills that require infrequent updates~\cite{ito2000mechanisms}. In contrast, the Cerebrum continuously adapts its strategies based on the nuances of new tasks and environments~\cite{demarin2014neuroplasticity,hallett2005neuroplasticity}. This biological parallel motivates our core hypothesis: an effective agent should pair a stable, proficient grounding model with a dynamic planner that is specialized for different software domains through targeted, experience-driven learning.

To realize this vision, we propose a trainable compositional framework that integrates Qwen2.5-VL~\cite{bai2025qwen2} as the planner (cerebrum) and UI-Tars-1.5~\cite{qin2025uitars} as the executor (cerebellum). Unlike prompting-based systems that rely on proprietary closed-source planners, our framework makes the planner itself learnable through interaction with software environments mediated by a static executor. Concretely, the executor provides stable, software-agnostic grounding for low-level GUI actions, while the planner, by leveraging this reliable interface, can gradually acquire domain-specific knowledge and improve its high-level planning strategies. In contrast to end-to-end training of a full agent, which requires massive amounts of specialized data and costly retraining of both perception and execution modules, our decoupled approach is substantially more data efficient: only the planner is optimized for domain adaptation, while the executor remains fixed as a general-purpose grounder that already possesses strong generalization ability after massive pretraining for grounding purposes. This design reduces reliance on curated trajectories, lowers training cost, and ensures controllable adaptation.

To train the planner effectively under this cerebrum–cerebellum separation, we avoid the need for costly human-labeled trajectories. Instead, we leverage a judging system built from open-source models to automatically provide dense reward signals, combined with autonomous interaction with scientific software environments through the static executor. This setup enables the planner to gradually acquire domain-specific planning ability with zero human effort. Furthermore, by distributing the interaction process across multiple software environments in parallel—coordinated by a central master—we can significantly accelerate reinforcement learning. This strategy not only makes the training process more efficient but also echoes our brain-inspired design: the cerebellum-like executor delivers stable grounding, while the cerebrum-like planner continually adapts through experience.

We validate our framework on four typical scientific software applications from the ScienceBoard benchmark~\citep{sun2025scienceboard}. Experiments show that our method not only significantly improves the baseline performance (Cerebrum: Qwen2.5-32B-VL, Cerebellum: UI-Tars-1.5) but also establishes a new state-of-the-art (SOTA) among open-source models, confirming its effectiveness.

\section{Related Works}
\noindent\textbf{Reinforcement Learning for LVLMs.}
Training for LLMs and LVLMs~\citep{touvron2023llama,grattafiori2024llama,liu2023visual,bai2025qwen2,wang2024qwen2,xing2025scalecap,sun2024bootstrap3d,sun2024xpromptuniversalincontextimage,ding2025mm} has progressed from data-intensive Supervised Fine-Tuning (SFT)~\citep{liu2023visual,wei2022chain} towards Reinforcement Learning (RL). Algorithms like Group Relative Policy Optimization (GRPO)~\citep{guo2025deepseek,shao2024deepseekmath} have proven effective for reasoning tasks, moving beyond earlier single-turn RLHF applications~\citep{ouyang2022training,ziegler2019fine,rafailov2023direct}. However, applying RL to complex agentic tasks~\citep{bai2024digirl,qi2024webrl,zhou2024archer,zhai2024fine,carta2023grounding} is challenging. Prevailing methods train monolithic agents end-to-end, often requiring co-trained critic models~\citep{schulman2015high} or preference-based optimization like DPO~\citep{rafailov2023direct,putta2024agent,qin2025uitars}, which problematically entangles the distinct skills of planning and execution. 
In contrast, our work employs a decoupled reinforcement learning strategy: the high-level planner is optimized via environmental interaction while the execution model remains fixed. We adapt GRPO by computing rewards from the final action and backpropagating the advantage exclusively through planning tokens. This targeted optimization stably enhances strategic planning, distinguishing our method from prior works that train dedicated critic models~\citep{bai2024digirl,qi2024webrl} or use filtered behavior cloning~\citep{pan2024autonomous,chen2020bail}.

\noindent\textbf{Computer Use Agent.}
Fueled by advancements in Large Vision-Language Models (LVLMs)~\citep{touvron2023llama,grattafiori2024llama,liu2023visual,bai2025qwen2,wang2024qwen2}, a new generation of agents capable of operating computers via multi-modal inputs is emerging~\citep{hu2024agents,hong2024cogagent,cheng2024seeclick,nguyen2024gui,lin2024showui,sun2024genesis}. Whether processing structured text and code~\citep{qi2024webrl,putta2024agent,lai2024autowebglm,code2024survey,nakano2021webgpt} or screenshots~\citep{hong2023cogagent,lin2024showui,wu2024atlas,operator}, these agents face an inherent dichotomy analogous to human cognition: the tension between high-level strategic planning and precise, low-level action execution.
This has motivated the development of compositional frameworks that decouple these responsibilities~\citep{agashe2024agent,agashe2025agent,liu2023bolaa,zhang2025appagent,song2025coact}. However, a significant portion of this research relies on static, non-trainable systems that orchestrate powerful, often proprietary models~\citep{claude,operator,google2025gemini25preview,yan2023gpt,he2024webvoyager,zhang2024android,wang2023voyager,wu2024copilot} as their core planner. This design fundamentally prevents the agent from adapting through experience---a critical flaw for mastering novel software where interaction data is scarce. Our work charts a different course by exploring reinforcement fine-tuning of the planner. By enabling the planner to learn specialized domain knowledge through direct software interaction via a fixed execution model, our strategy achieves robust performance on unfamiliar applications.
\section{Method}

\begin{figure}[t]
    \centering
    \includegraphics[width=\textwidth]{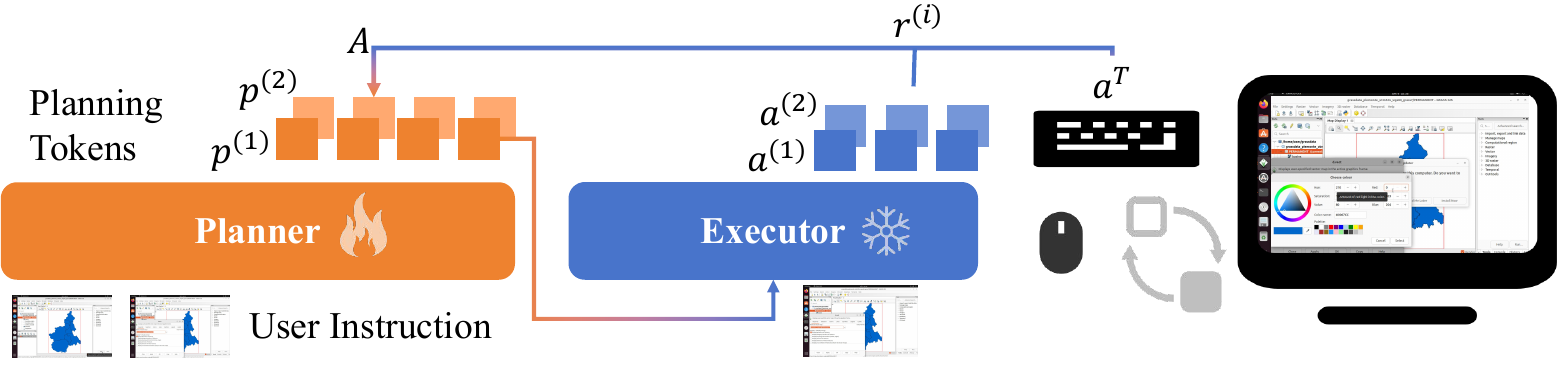}
    \caption{\textbf{Overall training process of the proposed Planner–Executor framework.} 
    The Planner generates high-level thoughts based on the history and screenshots, 
    while the Executor executes concrete GUI actions accordingly. During training, the rewards are calculated from $a^{(i)}$ and applied to $p^{(i)}$ to calculate loss. 
    }
    \label{fig:framework}
\end{figure}

\subsection{Problem Formulation}
We formally define the task of autonomous GUI operation for software workflows as a Partially Observable Markov Decision Process (POMDP). Each task is initiated with a natural language instruction $g$ from the task space $\mathcal{G}$. 
At each timestep $t$, the agent perceives the latent environment state $s_t \in \mathcal{S}$ through a visual observation $o_t \in \Omega$, consisting of a screenshot of the user interface.
The agent's behavior is governed by a policy $\pi$, instantiated by a large vision-language model, which synthesizes an action program $a_t \in \mathcal{A}$. 
The action space $\mathcal{A}$ consists of precisely parameterized \texttt{pyautogui} scripts, where precision in arguments (e.g., coordinates) is critical for execution. 
The policy generates this action based on the initial instruction and the history of interactions:
$$a_t = \pi(g, (o_1, a_1, \dots, a_{t-1}, o_t))$$
This sequential process induces a state trajectory $\tau = (s_0, s_1, \dots, s_T)$ with the maximum time step $T$. A task is considered successful if the final state $s_T$ satisfies the predefined goal condition specified in $\mathcal{G}$.

\subsection{Model Architecture}
To address the inherent trade-off in monolithic models, which struggle to balance long-horizon planning with precise action grounding, we propose a composite agent architecture that structures the decision-making process into a Planner-Executor framework. 
This design decouples the task into two distinct yet collaborative modules: a high-level Planner responsible for strategic thinking and a low-level Executor for concrete action execution.

\paragraph{Planner} The Planner is instantiated from the Qwen2.5-VL~\citep{bai2025qwen2} model. Its primary responsibility is to analyze the task's progress and formulate a high-level, explicit plan $p_t$ for each step. Specifically, at each timestep $t$, the Planner receives the interaction history up to the previous step $m_{t-1} = (p_1, a_1, \dots, p_{t-1}, a_{t-1})$, the current visual observation $o_t$, and the preceding observation $o_{t-1}$. The output is a structured thought, denoted as $p_t$, which outlines the immediate objective and explicitly identifies the target UI elements for interaction. The process can be summarized as:
$$p_t = \text{Planner}(m_{t-1}, o_{t-1}, o_t)$$

\paragraph{Executor} The Executor employs a UI-TARS-1.5~\citep{qin2025uitars} model. Its role is to translate the Planner's abstract thought $p_t$ into a precise, executable action. The Executor is provided with the same historical and visual context as the Planner ($m_{t-1}$, $o_{t-1}$, and $o_t$), but is critically augmented with the Planner's newly generated thought $p_t$. Its output is a low-level GUI action $a_t$ in the form of a `pyautogui' command, such as `click(x, y)'. The Executor's operation is defined as:
$$a_t = \text{Executor}(m_{t-1}, o_{t-1}, o_{t}, p_t)$$

\subsection{Training Pipeline}

Our training methodology employs a two-stage curriculum designed for initial specialization followed by broad generalization. 

\subsubsection{Stage 1: Specialization via Decoupled Reinforcement Learning}
\label{sec:decoupled_reinforcement_learning}
The primary objective of this initial training stage is to enhance the agent's specialized performance on individual software applications.

Through empirical analysis, we observed that the Executor exhibits strong generalization capabilities, accurately translating well-structured plans into executable actions. However, the Planner module emerged as the primary bottleneck, often struggling to formulate effective high-level strategies. To address this, we adopt a decoupled training strategy that focuses reinforcement learning exclusively on the Planner ($\pi_\theta = \text{Planner}$). This targeted approach allows us to refine the agent's strategic reasoning without altering the already competent Executor.

Since the initial Planner is relatively weak and generates a limited number of successful trajectories, standard reinforcement learning methods can be inefficient. Therefore, we adapt the Group Relative Policy Optimization (GRPO) framework \citep{guo2025deepseek,shao2024deepseekmath}, which is particularly effective in such scenarios. GRPO can derive a meaningful learning signal by comparing the relative quality of different outputs, even when most of them are suboptimal.

The training process for a given task unfolds as follows. Given the current state and interaction history, the Planner first generates a group of $G$ candidate plans. Subsequently, the fixed Executor takes each plan as input and produces a corresponding low-level action. To generate a fine-grained learning signal, we compute a reward for each plan by comparing its resulting action $a^{(i)}$ to the labeled positive action $a_T$ (details of labeling process are in Sec.\ref{sec:judge_system_methods} ). Our composite reward function assesses both the correctness of the action type and the precision of its parameters:

\begin{equation}
r^{(i)} = r(a^{(i)}, a_T) = \mathbb{I}\left(\text{type}(a^{(i)}) = \text{type}(a_T)\right) + r_{\text{dist}}(a^{(i)}, a_T),
\end{equation}

Here, the indicator function $\mathbb{I}(\cdot)$ provides a binary reward for selecting the correct type of action (e.g., \texttt{click} vs. \texttt{type}). The term $r_{\text{dist}}(a^{(i)}, a_T)$ offers a continuous reward based on the parametric similarity between the predicted and ground-truth actions, such as L1 distance for coordinates or IoU for bounding boxes. These distance-based rewards are normalized to $[0, 1]$ to ensure consistent scaling. 

Once the rewards are calculated, they are used to derive a relative advantage $A^{(i)}$ for each plan, which is then fed into the GRPO loss function to update the Planner policy:

\begin{equation}
A^{(i)} = \frac{r^{(i)} - \text{mean}(\{r^{(j)}\}_{j=1}^{G})}{\text{std}(\{r^{(j)}\}_{j=1}^{G})}, \quad i=1,\cdots,G.
\end{equation}

The GRPO loss is formulated as follows:

\begin{align}
\mathcal{L}_{\text{GRPO}}(\pi_\theta) &=
- \mathbb{E}_{(s, I) \sim \mathcal{D}, \{a^{(i)}\}_{i=1}^G \sim \pi_{\text{ref}}(\cdot \mid s, I)}
\\\Bigg[ \frac{1}{G} \sum_{i=1}^G \frac{1}{|p^{(i)}|}& \sum_{t=1}^{|p^{(i)}|}
\Big\{
\min\Big(
r_t^{(i)}(\theta) A^{(i)},
\text{clip}(r_t^{(i)}(\theta), 1 - \epsilon, 1 + \epsilon) A^{(i)}
\Big)
- \beta \, D_{\text{KL}}^{(i,t)}(\pi_\theta \| \pi_{\text{ref}})
\Big\}\Bigg],\nonumber
\end{align}

\begin{align*}
    \text{where} \quad
    r^{i,t}(\theta) = \frac{\pi_{\theta}(p^{(i)}|s, I)}{\pi_{\theta_\text{ref}}(p^{(i)}|s, I)}
    \quad \text{and} \quad
    D_\text{KL}^{i,t}(\pi_\theta, \pi_\text{ref})=
    \frac{\pi_\text{ref}(p^{(i)}|s, I)}{\pi_{\theta}(p^{(i)}|s, I)} - 1 - \log \frac{\pi_\text{ref}(p^{(i)}|s, I)}{\pi_{\theta}(p^{(i)}|s, I)}.
\end{align*}

Consistent with the approach in \citep{shao2024deepseekmath,guo2025deepseek}, this advantage is applied across all reasoning tokens in the plan $p^{(i)}$, encouraging the model to develop more robust and free-form planning capabilities.

\subsubsection{Stage 2: Generalization via Aggregated Supervised Fine-Tuning}

We adopt the specialist-to-generalist paradigm proposed in~\cite{sun2025seagent}, where a generalist model is trained by leveraging multiple specialist models as teachers. We observe that directly applying reinforcement learning across all software leads to suboptimal performance. To address this, we first train four specialist models using the methods described in Sec.~\ref{sec:decoupled_reinforcement_learning}. These specialists are then employed to generate new trajectories for each software, which serve as supervision for training a generalist model. After learning from the four software-specific teachers, the resulting generalist not only surpasses its teachers in performance, but also demonstrates stronger reasoning and reflection abilities during planning, as well as broader domain knowledge across different software. 

\subsection{Auto Exploration Pipeline.}

\textbf{Auto Task Generation.}
\label{sec:auto_task_generation}
We employ Qwen2.5-72B~\citep{wang2024qwen2} as the task generator to produce high-level tasks. Specifically, a small set of real human-instructed tasks on each software is provided as input, together with the prompt shown in Fig.~\ref{fig:task_generation_prompt}. The agent then repeatedly executes these tasks to collect a diverse set of interaction trajectories, which are subsequently filtered by a judge system to retain only trajectories with positive actions for training.

\textbf{Judge System for Providing Reward Signals.}
\label{sec:judge_system_methods}
Our judge system labels the positive actions $a_T$ within an agent’s trajectory when performing a task. Given a full trajectory $\mathcal{H} = \{o_0, a_0, \dots, o_{\text{final}}\}$, the judge takes the complete sequence of screenshot observations $(o_1, o_2, \dots, o_n)$ as input and outputs three signals: \texttt{Correctness}, \texttt{Redundant}, and \texttt{First Error Step}, using the detailed prompt shown in Fig.~\ref{fig:judge_prompt}. A trajectory is considered clean and successful when \texttt{Correctness} is \texttt{True} and both \texttt{Redundant} and \texttt{First Error Step} are empty. In this case, all actions $a$ in the trajectory are labeled as $a_T$. We present a detailed evaluation of the judge’s precision and discuss approaches for improving it in Sec.~\ref{sec:judge_eval}.

\begin{figure}[htb]
    \centering
    \begin{subfigure}{0.65\textwidth}
        \centering
        \includegraphics[width=\textwidth]{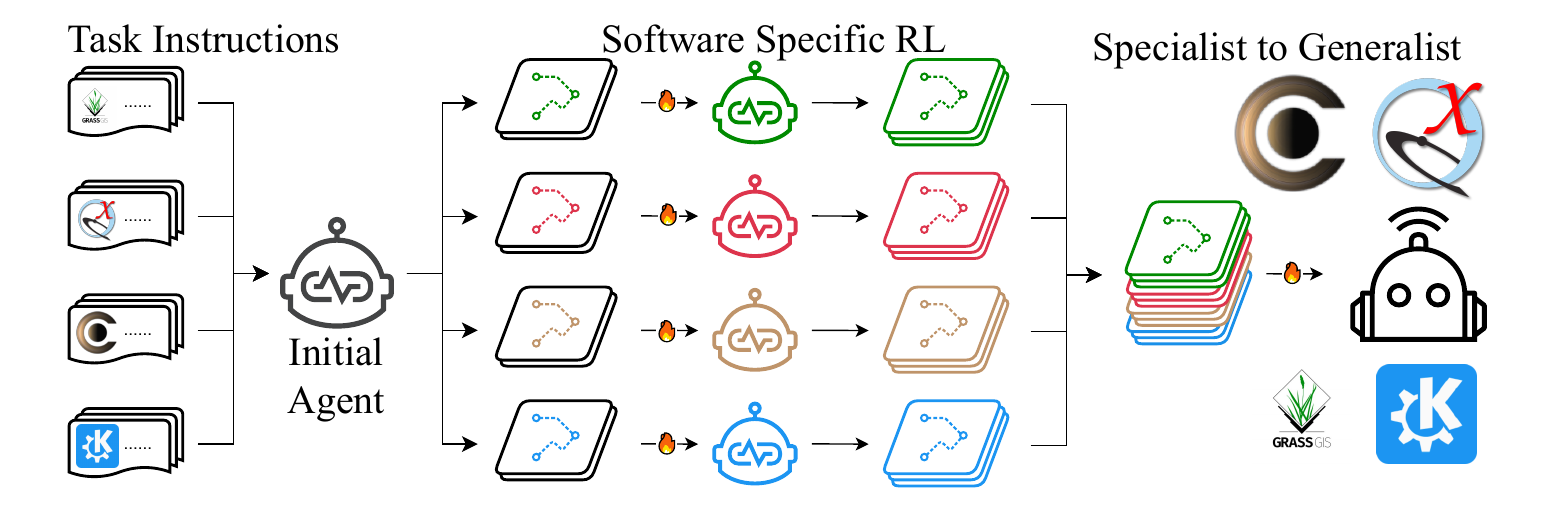} 
        \caption{Specialist-to-Generalist strategy.}
        \label{fig:sub1}
    \end{subfigure}
    \hfill
    \begin{subfigure}{0.32\textwidth}
        \centering
        \includegraphics[width=\textwidth]{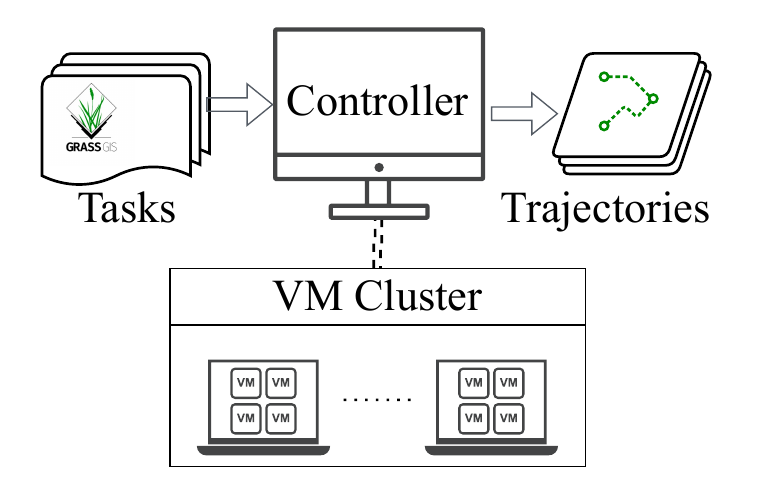}
        \caption{Distributed VM System.}
        \label{fig:sub2}
    \end{subfigure}
    \caption{\textbf{Exploration pipeline for training support.} } 
    \label{fig:main}
\end{figure}

\textbf{Distributed Virtual Machine System.} Task execution is the most time-consuming step in our pipeline, so we developed a lightweight distributed system to accelerate large-scale trajectory curation. As illustrated in Fig.~\ref{fig:sub2}, the system follows an HTTP-based master–client architecture: the master node manages a dynamic task queue, monitors execution progress, and aggregates results, while multiple client nodes execute tasks in parallel within isolated virtual machine environments. This design enables efficient scaling to hundreds of concurrent environments, substantially reducing the time required to collect successful trajectories and making the framework well-suited for large-scale training and evaluation.

\section{Experiments}

\begin{figure}[t]
    \centering
    \includegraphics[width=\textwidth]{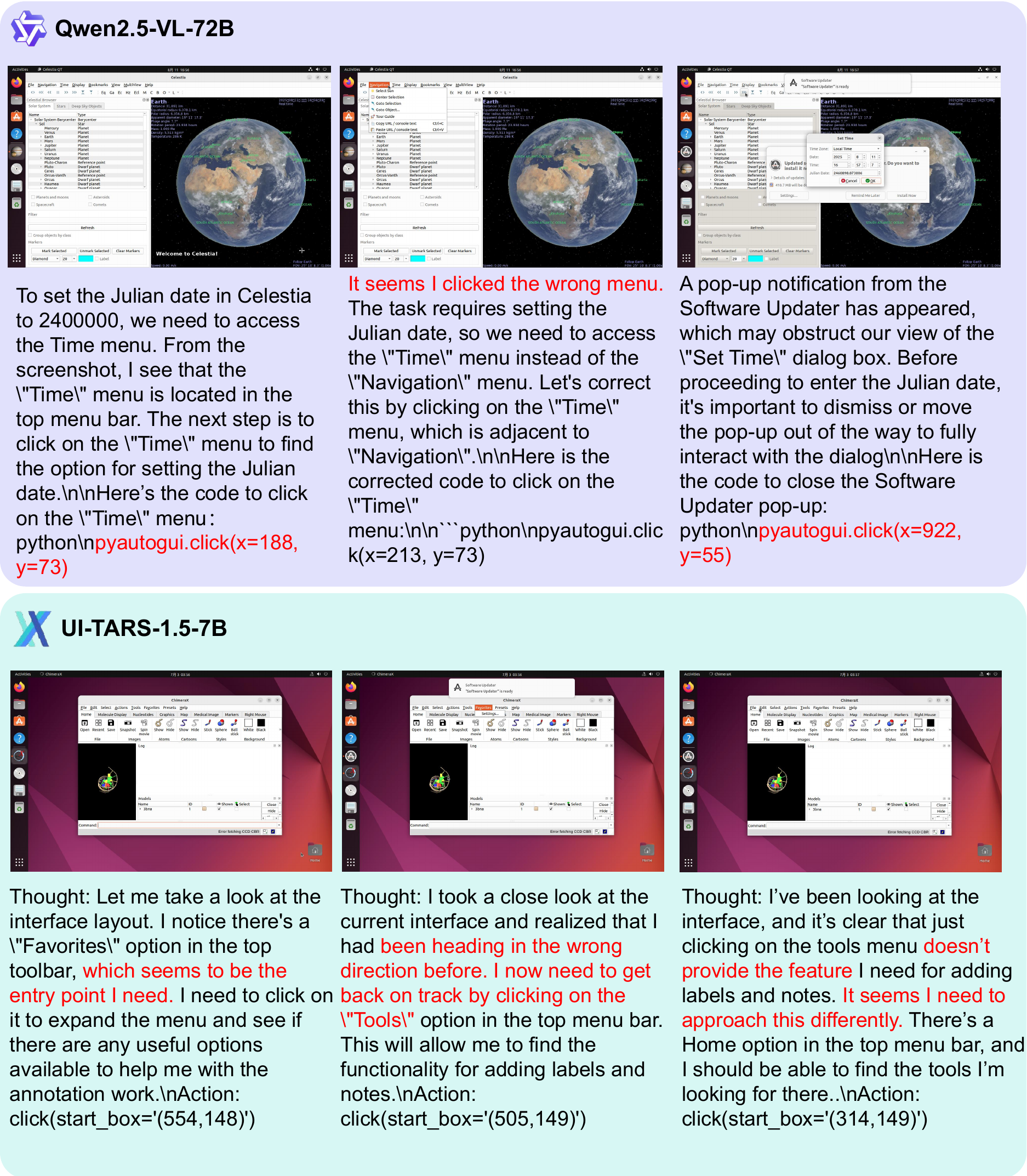}
    \caption{\textbf{Case studies.} Qwen2.5-VL-72B~\cite{bai2025qwen2} struggles with precise grounding, whereas UI-TARS-1.5~\cite{qin2025uitars}, though specialized, fails to generalize to out-of-distribution software.}
    \label{fig:framework}
\end{figure}

\begin{table}[h!]
\centering
\scalebox{0.82}{
\begin{tabular}{clccccc}
\toprule
\multirow{2}{*}{\textbf{Metrics}} & \multirow{2}{*}{\textbf{Model}} & \multicolumn{5}{c}{\textbf{Success Rate ($\uparrow$)}} \\
\cmidrule(lr){3-7}
& & \textbf{Algebra} & \textbf{Biochem} & \textbf{GIS} & \textbf{Astron} & \textbf{Overall} \\
\midrule
\multirow{9}{*}{Average@1} 
& \cellcolor{lightpurple}GPT-4o~\citep{gpt4} & 3.23\% & 0.00\% & 0.00\% & 0.00\% & 0.81\% \\
& \cellcolor{lightpurple}Claude-3.7-Sonnet~\citep{claude-3-7} & 9.67\% & 37.93\% & 2.94\% & 6.06\% & 14.15\% \\
& \cellcolor{lightpurple}Gemini-2.0-Flash~\citep{team2023gemini} & 6.45\% & 3.45\% & 2.94\% & 6.06\% & 4.73\% \\
& \cellcolor{lightpurple}GPT4o$\xrightarrow{}$UGround-V1-7B~\citep{gou2024navigating} & 0.00\% & 3.45\% & 0.00\% & 3.03\% & 1.62\% \\
& \cellcolor{lightpurple}GPT4o$\xrightarrow{}$OS-Atlas-Pro-7B~\citep{wu2024atlas} & 6.25\% & 10.34\% & 0.0\% & 3.03\% & 4.92\% \\
& \cellcolor{lightpurple}GPT4o$\xrightarrow{}$UI-TARS-72B~\citep{qin2025uitars} & 3.23\% & 10.34\% & 5.88\% & 6.06\% & 6.38\% \\
& \cellcolor{lightgreen}Qwen2.5-VL-72B~\citep{bai2025qwen2} & 22.58\% & 27.59\% & 5.88\% & 9.09\% & 12.94\% \\
& \cellcolor{lightgreen}InternVL3-78B~\citep{zhu2025internvl3} & 6.45\% & 3.45\% & 0.00\% & 0.00\% & 2.69\% \\
& \cellcolor{lightgreen}UI-TARS-1.5-7B~\citep{qin2025uitars} & 12.90\% & 13.79\% & 0.00\% & 6.06\% & 8.19\% \\
\midrule
\multirow{4}{*}{Average@8} 
& \cellcolor{lightgreen}Qwen2.5-VL-32B~\citep{bai2025qwen2} & 10.48\% & 13.79\% & 1.47\% & 4.55\% & 7.57\% \\
& \cellcolor{lightgreen}UI-TARS-1.5-7B~\citep{qin2025uitars} & 6.49\% & 10.24\% & 0.80\% & 3.03\% & 5.14\% \\
& \cellcolor{lightgreen}\textbf{\methodname(Stage-1)}* & 13.71\% & 26.29\% & 7.72\% & 9.85\% & 14.39\% \\
& \cellcolor{lightgreen}\textbf{\methodname(Stage-2)} & \textbf{20.16\%} & \textbf{32.23\%} & \textbf{14.71\%} & \textbf{17.05\%} & \textbf{21.04\%} \\
\midrule
\multirow{4}{*}{Pass@8} 
& \cellcolor{lightgreen}Qwen2.5-VL-32B~\citep{bai2025qwen2} & 29.03\% & 31.03\% & 8.82\% & 9.09\% & 19.49\% \\
& \cellcolor{lightgreen}UI-TARS-1.5-7B~\citep{qin2025uitars} & 19.35\% & 24.14\% & 5.88\% & 12.12\% & 15.36\% \\
& \cellcolor{lightgreen}\textbf{\methodname(Stage-1)}* & 41.94\% & 44.83\% & 23.53\% & 18.18\% & 32.12\% \\
& \cellcolor{lightgreen}\textbf{\methodname(Stage-2)} & \textbf{48.39\%} & \textbf{51.72\%} & \textbf{29.41\%} & \textbf{30.30\%} & \textbf{39.96\%} \\
\bottomrule
\end{tabular}
}
\caption{Success rates of various models on ScienceBoard~\citep{sun2025scienceboard}. 
\colorbox{lightpurple}{Proprietary models} 
and 
\colorbox{lightgreen}{open-sourced models} based methods
are highlighted with purple and green backgrounds, respectively. *Indicates specialist agents trained separately for each software with ensembled results.}
\label{tab:scienceboard_result}
\end{table}

\subsection{Agent Performance Evaluation.}
Our planner-executor approach is based on Qwen2.5VL-32B~\citep{bai2025qwen2} serve as planner and UI-TARS-1.5-7B~\citep{qin2025uitars} serve as executor. We use method proposed in Sec.\ref{sec:auto_task_generation} to generate high level tasks for each software from ScienceBoard~\citep{sun2025scienceboard}. 
through decoupled reinforcement learning proposed in Sec.\ref{sec:decoupled_reinforcement_learning}. During Training, the reward signal is provided by our judge system evaluated in Sec.\ref{sec:judge_eval}. Our training is based on OpenRLHF~\cite{hu2024openrlhf}. As reported in Tab.\ref{tab:scienceboard_result}, our evaluation is done on four GUI centric software from ScienceBoard~\cite{sun2025scienceboard}. We also report other planner-executor decoupled approaches. This first-stage reinforcement learning approach lead to significant performance gain compared to baseline. 

In second stage, we use these specialist planner serve as teachers to teach a generalist planner. This new model is also initialized from Qwen2.5VL-32B and perform supervised fine-tuning on 0.77K trajectories from teacher models labeled by our judge system. As shown in tab.\ref{tab:scienceboard_result}, this new model surpass the performance of the ensemble of individual specialist, showing improved reasoning and planning abilities. This result demonstrates the effectiveness of our specialist-to-generalist strategy. 

\begin{table}[t]
    \centering
    \small
    \caption{Evaluation of different judge methods on AgentRewardBench~\citep{lu2025agentrewardbench} and ScienceBoard~\citep{sun2025scienceboard}.}
    \label{tab:judge_eval}
    \begin{tabular}{lcccc}
        \toprule
        \multirow{2}{*}{Method} & \multicolumn{2}{c}{AgentRewardBench~\citep{lu2025agentrewardbench}} & \multicolumn{2}{c}{ScienceBoard~\citep{sun2025scienceboard}} \\
        \cmidrule(lr){2-3} \cmidrule(lr){4-5}
                                & Precision & Recall & Precision & Recall \\
        \midrule
        Qwen2.5-VL-72B-single              & 64.5      & 83.4   & 41.5      & 80.1   \\
        72B-GUI-Judge                      & 73.5      & 79.0   & 43.7      & 80.1   \\
        72B-voting@4                       & 76.1      & 79.5   & 58.6      & 75.3   \\
        72B-voting@4 w/ multi-res          & 78.9      & 77.4   & 65.7      & 77.9   \\
        72B-voting@4 Ensemble              & 81.2      & 76.8   & 69.5      & 74.2   \\
        \bottomrule
    \end{tabular}
\end{table}

\subsection{Towards Precise Judging System}
\label{sec:judge_eval}
Our reinforcement learning framework heavily relies on accurate judgments of agent trajectories to provide reliable reward signals. In this section, we present a detailed evaluation of our judge model, which demonstrates improved precision in decision making.

\textbf{Settings.} We conduct experiments on two sources of trajectories. (1) AgentRewardBench~\citep{lu2025agentrewardbench}, a benchmark designed specifically for judge evaluation. (2) A trajectory dataset we collected from ScienceBoard~\citep{sun2025scienceboard}. We run Qwen2.5-VL-72B~\citep{bai2025qwen2} on ScienceBoard tasks and extract 377 labeled trajectories, which are then used as inputs to our judge model. This setup allows us to quantitatively assess the judge’s ability to discriminate between successful and failed executions. We report \emph{Precision} and \emph{Recall} as our primary metrics. For voting-based strategies, we adopt a sampling temperature of $T=1.0$ and a nucleus sampling probability of $top\_p=0.6$ over 4 independent inference runs.

\textbf{Results.} As summarized in Table~\ref{tab:judge_eval}, our evaluations reveal three effective strategies for improving precision, building upon difference description fine-tuning~\citep{sun2025seagent}:
1. \emph{Voting.} Instead of a single query, we prompt the model multiple times with high randomness ($T=1.0$, $top\_p=0.6$). A trajectory is only deemed successful if all votes agree, which significantly reduces false positives.  
2. \emph{Multi-resolution inputs.} Trajectories often include long sequences of high-resolution screenshots. We observe that using a mixture of resolutions across voting rounds is beneficial: low-resolution images help capture global execution dynamics, while high-resolution images aid in detecting fine-grained correctness. In practice, we first apply low-resolution inputs to quickly filter out failures, thereby improving both precision and efficiency. 
3. \emph{Model ensembling.} In addition to the fine-tuned judge model (see Sup.~\ref{sup:judge_model_detail}), we find that ensembling two models within the voting strategy further enhances precision.

Across both ScienceBoard~\cite{sun2025scienceboard} and AgentRewardBench~\cite{lu2025agentrewardbench}, we observe a consistent progression: the fine-tuned model (72B-GUI-Judge) primarily improves recall, while voting substantially increases precision; multi-resolution inputs add further gains, and ensembling achieves the best balance with the highest precision while maintaining competitive recall. This consistent trend across benchmarks highlights the robustness and generality of our proposed strategies. With methods proposed in . This judge system provide high quality reward signal for the planner to perform RL to improve reasoning ability and learning software domain knowledge.



\section{Conclusion}
We presented a trainable Planner–Executor disentangled framework for GUI agents, inspired by the division of labor between the cerebrum and cerebellum. By coupling a fixed executor (UI-Tars-1.5) with a fine-tunable planner (Qwen2.5-VL), and supporting it with a robust judging system, GRPO-based exploration, and a distributed data generation pipeline, our approach effectively addresses the challenges of complex interfaces and long-horizon planning. Experiments on ScienceBoard applications demonstrate substantial improvements over strong baselines, establishing a new open-source state-of-the-art. These results highlight the importance of combining stable execution with adaptive planning, and open promising directions for extending our framework to richer multi-modal feedback, broader professional domains, and continual learning for long-term adaptability.

\bibliography{iclr2025_conference}
\bibliographystyle{iclr2025_conference}

\newpage
\appendix
\section{Judge Model Fine-tuning Details}
\label{sup:judge_model_detail}

Inspired by~\cite{sun2025seagent}, we adopt a fine-tuning approach to obtain a strong judge model. We scale up the model to Qwen2.5-VL-72B~\cite{bai2025qwen2}, and use a dataset comprising 4.7K labeled judgment samples. These trajectories are generated by Qwen2.5-VL and Gemini-2.0-Pro on WebArena~\cite{zhou2023webarena}, UI-TARS-1.5~\cite{qin2025uitars}, and GPT-4o~\cite{gpt4} on OSWorld~\cite{qin2025uitars}. Judgments are provided by GPT-4o and Gemini2.5-Pro~\cite{google2025gemini25preview}, with detailed captions for each screenshot frame during agent execution. The judgments are further filtered, retaining only those that align with verified ground-truth results. Additionally, change description data is incorporated inspired by SEAgent~\cite{sun2025seagent}.

Training is conducted on 32 A100 GPUs for 370 steps, using LoRA~\cite{hu2022lora} with a rank of 8. The resulting model, trained on OSWorld trajectories, generalizes well to AgentRewardBench~\cite{lu2025agentrewardbench} and ScienceBoard~\cite{sun2025scienceboard}. This fine-tuned model is referred to as \textbf{72B-GUI-Judge} in Table~\ref{tab:judge_eval}, and demonstrates improved precision on two out-of-domain benchmarks. When further ensembled with the original 72B base model, it achieves even higher precision, providing more accurate reward signals—crucial for effective reinforcement learning of the planner agent.

\section{Prompt Details.}
\begin{figure}[t]
    \centering
    \includegraphics[width=\textwidth]{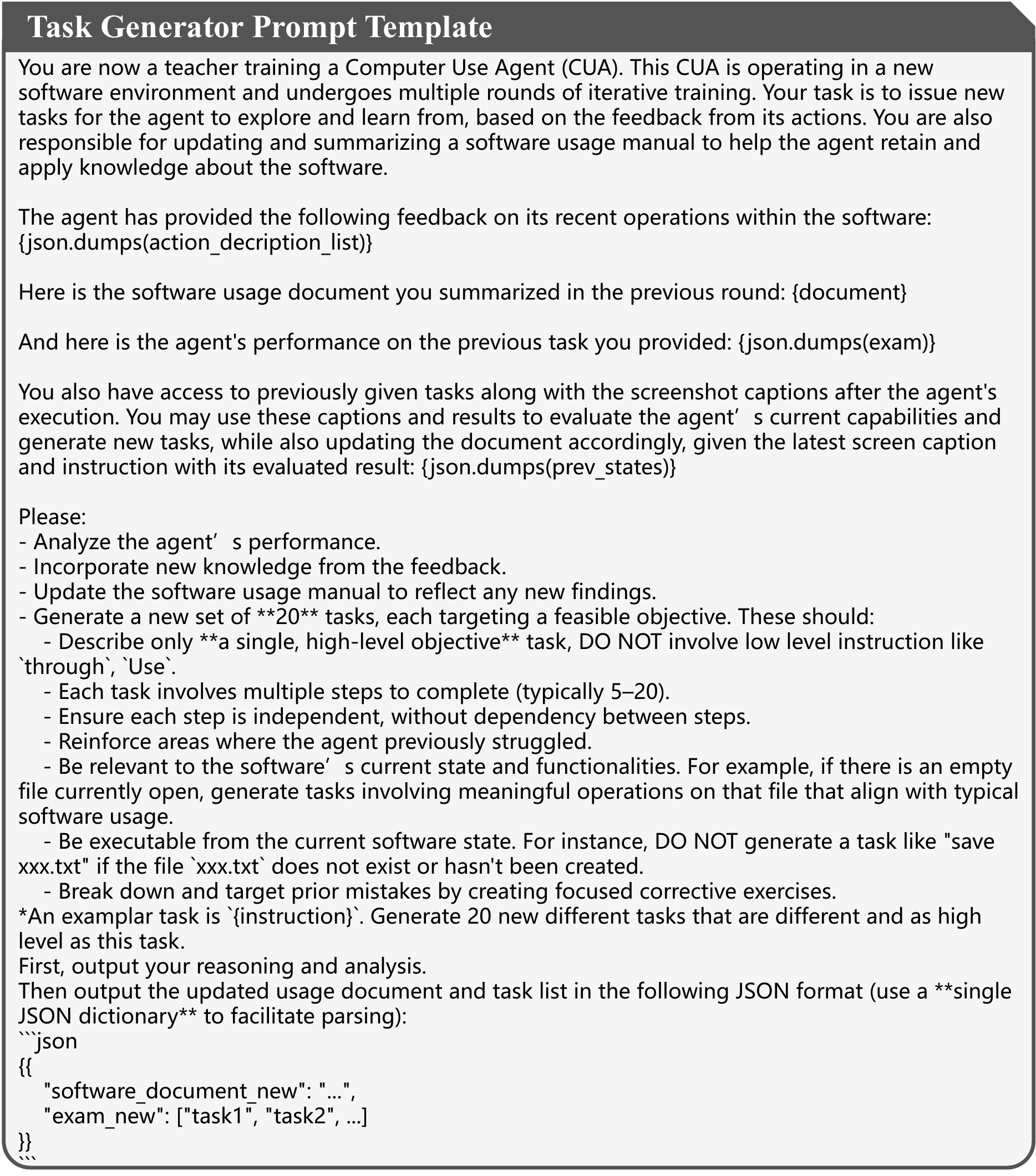}
    \caption{\textbf{Detailed prompt for task generation.}}
    \label{fig:task_generation_prompt}
\end{figure}

\begin{figure}[t]
    \centering
    \includegraphics[width=\textwidth]{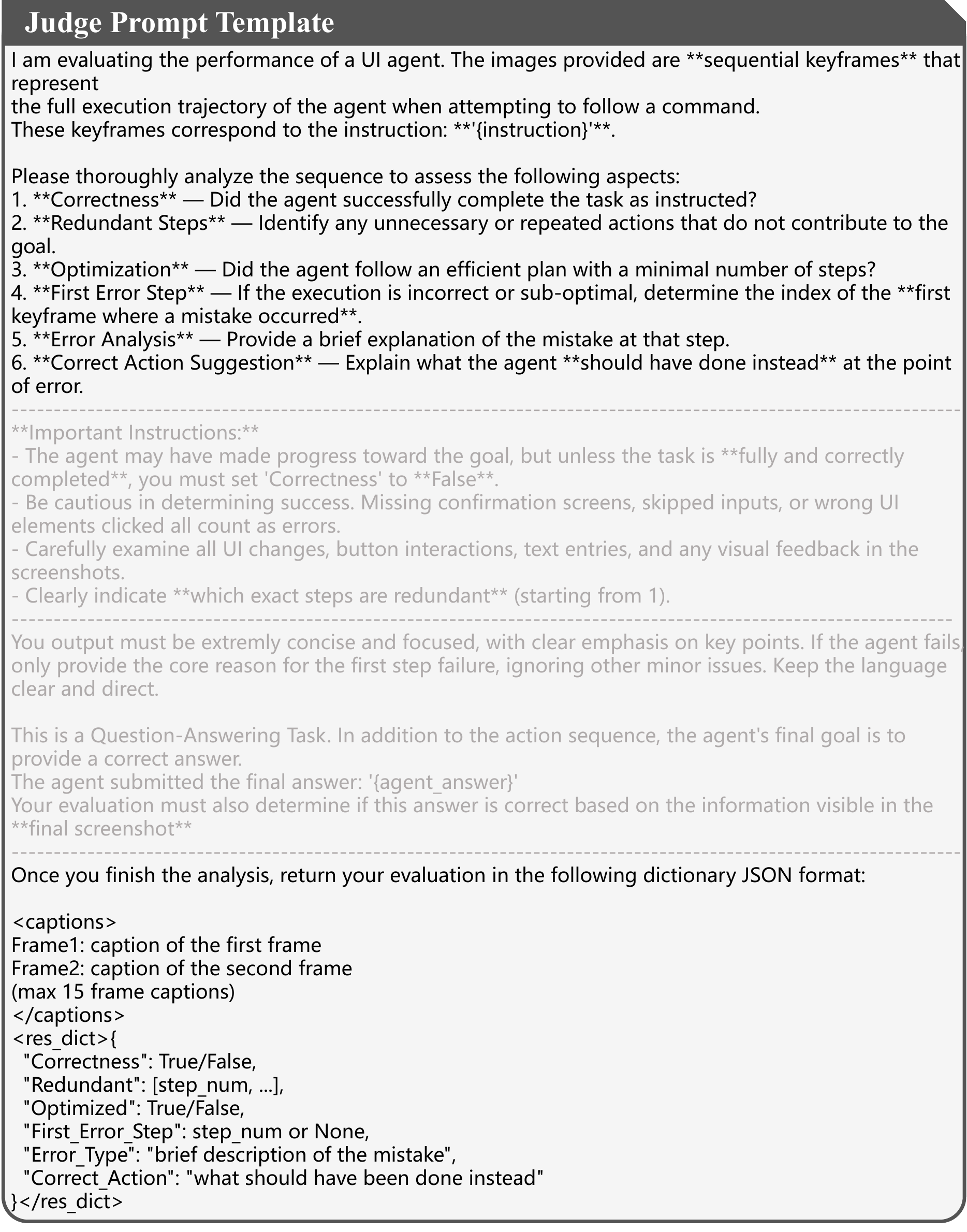}
    \caption{\textbf{Detailed prompt for the judge model.} Text in gray all task type based on whether finish given task or answer the question from user.}
    \label{fig:judge_prompt}
\end{figure}

\begin{figure}[t]
    \centering
    \includegraphics[width=\textwidth]{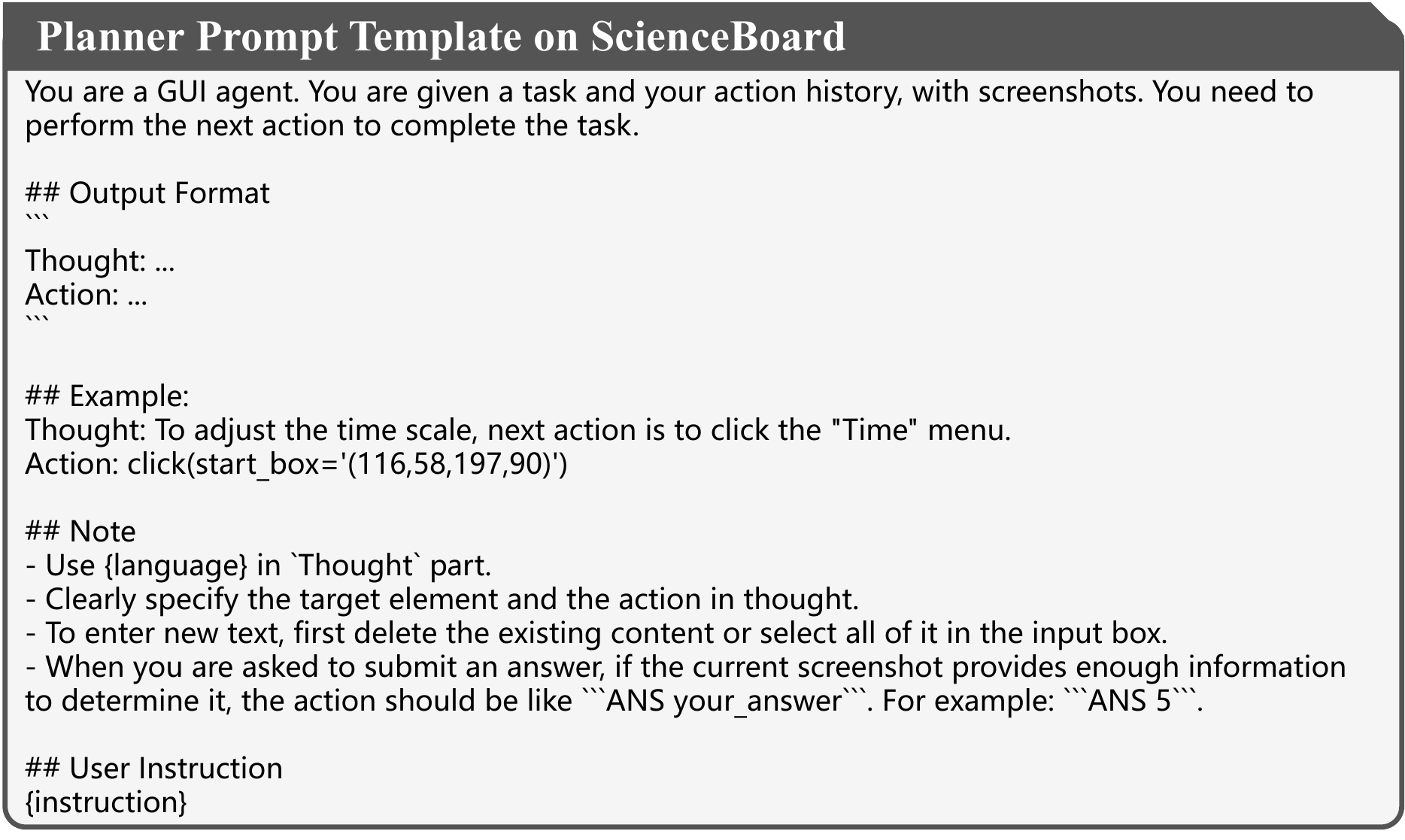}
    \caption{\textbf{Detailed prompt for the planner agent.}}
    \label{fig:agent_prompt}
\end{figure}

We provide detailed prompt for task generator in Fig.\ref{fig:task_generation_prompt} and judge system in Fig.\ref{fig:judge_prompt}. Detailed prompt for planner agent is in Fig.\ref{fig:agent_prompt}. Prompt we used for executor agent aligns with UI-TARS~\cite{qin2025uitars} official code. 

\section{Virtual Machine System Details} We utilized a local cluster consisting of 15 servers to collect interaction trajectories. Among these, 13 servers were equipped with AMD EPYC 7742 processors, and 2 servers were equipped with Intel i9-13900K CPUs paired with NVIDIA GeForce RTX 4090 GPUs to support software with high graphical computing demands, such as ChimeraX. Using VMware Workstation Pro, we ran 4 to 8 independent virtual machines concurrently on each server to execute tasks in parallel.

\end{document}